%% file: main.tex
\documentclass[preprint,12pt,authoryear]{elsarticle}

\usepackage{amssymb}

\input{header}

\journal{Neural Networks SI: NN Learning in Big Data}

\begin{document}

\begin{frontmatter}



\title{Deep learning of fMRI big data: a novel approach to subject-transfer decoding}


\author[ku,atr]{Sotetsu Koyamada\corref{cor}}
\ead{koyamada-s@sys.i.kyoto-u.ac.jp}
\author[ku,atr]{Yumi Shikauchi}
\author[ku]{Ken Nakae}
 \author[ku]{Masanori Koyama}
\author[ku,atr]{Shin Ishii}
\address[ku]{Graduate School of Informatics, Kyoto University, Kyoto 606-8501, Japan}
\address[atr]{ATR Cognitive Mechanisms Laboratories, Kyoto 619-0288,
 Japan}
\cortext[cor]{Corresponding author}


\begin{abstract}
\input{abstract}
\end{abstract}


\end{frontmatter}


\input{full_article}



\bibliographystyle{elsarticle-harv}
\bibliography{bibliography/converted_to_latex.bib}






\end{document}

%% file: header.tex
\usepackage{amsmath}
\usepackage{natbib}
\usepackage{hyperref}
\usepackage{epstopdf}

\newcommand{\bvec}[1]{\mbox{\boldmath $#1$}}

\newcommand{\RR}{\mathbb{R}}



%% file: abstract.tex
As a technology to read brain states from measurable brain activities, \textit{brain decoding} are widely applied in industries and medical sciences.
In spite of high demands in these applications for a universal decoder that can be applied to all individuals simultaneously, large variation in brain activities across individuals has limited the scope of many studies to the development of individual-specific decoders.
In this study, we used deep neural network (DNN), a nonlinear hierarchical model, to construct a subject-transfer decoder.
Our decoder is the first successful DNN-based subject-transfer decoder.
When applied to a large-scale functional magnetic resonance imaging (fMRI) database, our DNN-based decoder achieved higher decoding accuracy than other baseline methods, including support vector machine (SVM).
In order to analyze the knowledge acquired by this decoder, we applied principal sensitivity analysis (PSA) to the decoder and visualized the discriminative features that are common to all subjects in the dataset.
Our PSA successfully visualized the subject-independent features contributing to the subject-transferability of the trained decoder.

%% file: keyword.tex
fMRI \sep
Brain decoding \sep
subject-transfer decoding \sep
deep neural network (DNN)\sep
principal sensitivity analysis (PSA)\sep
brain machine interface (BMI)

%% file: full_article.tex
\section{Introduction}
Brain decoding 
is an act of decoding exogenous and/or endogenous brain states from measurable brain activities \citep{haxby2001distributed,cox2003functional,kamitani2005decoding,shibata2011perceptual,horikawa2013neural}.
It has been attracting much attention in medical and industrial fields as a major next-generation technology.
Possible applications include brain machine interface (BMI) \citep{laconte2011decoding}, neuro rehabilitation \citep{sitaram2012acquired} and therapy of mental disorders \citep{sitaram2007fmri}.
Brain decoding is a function that takes brain activities as input and brain states as output, and its performance is evaluated by how well it approximates the real association between these two entities.
As such, it falls into the category of machine learning, and it is often studied in the particular framework of supervised learning \citep{lemm2011introduction}.
As in the case of any other applications of machine learning, the performance of the decoder depends heavily on the quality and quantity of the data used for its training.

Much difficulty remains in obtaining sufficient data from a single individual to build a reliable decoder.
One can extract only so much information from a single subject, because there is a limit to the mental and physical stress that the subject can endure.
One would therefore seek a decoder that can be trained using a big data amassed from multiple subjects, so that
we can reduce the stress per subject while granting the decoder an
ability to simultaneously decode heterogeneous dataset, i.e., a subject-transfer decoder \citep{fazli2009subject,raizada2012makes,marquand2014bayesian}.
For an ideal subject-transfer decoder, its decoding accuracy does not deteriorate over the dataset obtained from the population outside the group of individuals used in its training.
However, large subject-wide variation of the brain activities has long hindered the development of such decoders. 
The scope of the brain decoding studies up until recent years has hence been restricted to subject-specific (\textit{tailor-made}) decoder that can only cope with the data from the very subject who provided the training dataset (e.g., \citet{cox2003functional,haxby2001distributed,nishimoto2011reconstructing,horikawa2013neural}).

In this work, we will present a subject-transfer decoder in the form of a deep neural network (DNN) trained with big data, a decoder aimed at classifying the brain activities into seven cognitive task categories.
For both training and testing, we used a large fMRI dataset in Human Connectome Project (HCP) gathered from over 500 subjects.
The application of DNN to fMRI dataset is not new; \citet{plis2013deep} used a fMRI-trained DNN to study schizophrenia patients, and \citet{hatakeyama2014multi} used still another variation of fMRI-trained DNN to classify hand motions.
Our work is the first of its kind in using a DNN to construct a subject-transfer decoder from big data.
Our subject-transfer decoder achieved higher decoding accuracy than any other baseline methods like support vector machine (SVM).
This is indicative of DNN's superior generalization ability over heterogeneous big data.
Also, decoding accuracy improved monotonically as the number of training subjects increased.
In the light of the fact that we are engaged in subject-transfer decoding, this monotonic trend suggests that our training is successfully extracting more subject-independent features from larger dataset.
This also shows that the size of the dataset contributes to the robustness of the decoder over heterogeneous population.
DNN together with big data thus emerges as an successful new approach to the subject-transfer decoding.

In order to further assess the efficiency of our trained DNN, we applied principal sensitivity analysis (PSA) \citep{Koyamada2014}, a brand new knowledge discovery procedure, to highlight the subject independent features
used by the decoders for its function.
By illustrating these features on the map of brain, we were able to make some connections between these features and functional connectivity reported in human fMRI studies \citep{raichle2001default,raichle2007default,taylor2009two,cole2013multi}.

This paper is structured as follows.
In the method section, we will provide the settings under which we trained our DNN, along with the specification of the dataset.
We will also provide the theoretical background of the PSA.
In the result section, we will compare the performance of our DNN against standard classification techniques, and show how the decoding accuracy of the DNN improved as the size of the training dataset increased.
We will also provide an interpretation of the PSA in terms of functional connectivity in the brain.

\section{Methods}

\subsection{fMRI data acquisition and preprocessing}
Human Connectome Project (HCP) is a scientific project ``to map macroscopic human brain circuits and their relationship to behavior in a large population of healthy adults'' \citep{van2013wu}, and it provides one of the largest open databases of fMRI that are publically available today.
In this study, we used the task-evoked fMRI data collected from 499 participants in Quarter 1 through Quarter 6,
which were  preprocessed and registered by \citet{van2013wu,glasser2013minimal} (HCP S500 release).
For more details, see the HCP release reference manual\footnote{www.humanconnectome.org/documentation/S500}.

In this section, we will provide key data specifications and preprocessing procedure.
fMRI data of $499$ healthy adults were acquired by a Siemens 3T Skyra, with TR = $720$ ms, TE = $33.1$ ms, flip angle $52^\circ$, FOV = $208 \times 180$ mm, $72$ slices, $2.0 \times 2.0$ mm in plane resolution.
The preprocessing that had been applied to the fMRI data in the HCP prior to our own modification includes removal of spatial artifacts and distortions, within-subject cross-modal registrations, reduction of the bias field, and alignment to standard space.
In addition to these processes, we applied voxel-wise z-score transformation to the data and averaged the intensity over each anatomical region of interest (aROI).
The intention of the latter averaging procedure is to help the decoder learn features that are robust against
large inter-subject variability of brain activities.
aROIs were determined by the automated anatomical labeling method \citep{tzourio2002automated}.
In the end, the dimension per each preprocessed fMRI scan became 116.

The 499 participants (subjects) in the dataset we studied were asked to perform seven tasks related to the following categories: Emotion, Gambling, Language, Motor, Relational, Social and Working Memory (WM).
Each subject performed each task twice with time limits that varies across different tasks (see Table\,\ref{tab:1}).
Note that the number of scans conducted in the experiment varies across different tasks.
One hundred unrelated subjects completed all seven tasks.
The WM class occupied the largest proportion ($20.88$\%) of all scans for each subject.
The experimental design of each task is summarized below.
See \citet{barch2013function} for more details.
\begin{enumerate}
 \item \textbf{Emotion:} Participants were asked to match one of two simultaneously presented images with a target image (angry face or fearful face). This is a modified version of the emotion task employed in \citet{hariri2002amygdala}.
 \item \textbf{Gambling:} Participants were asked to play a simple game to get money. See \citet{delgado2000tracking} for more details.
 \item \textbf{Language:} After listening to a brief story, participants were asked to answer a two-alternative forced choice question about the topic of the story. See \citet{binder2011mapping} for more details.
 \item \textbf{Motor:} Participants were requested to move one of five body parts (left or right finger, left or right toe, or tongue) as instructed by a visual cue \citep{buckner2011organization}.
 \item \textbf{Relational:} Each participant was presented with two pairs of objects, and was subsequently asked to answer a second-order question regarding the shapes/textures of the objects.
 \item \textbf{Social:} Participants were presented with a movie clip, and were asked to decide whether the movements of the objects in the clips are related with each other in some way. The movie clips were originally prepared by \citet{castelli2000movement} and \citet{wheatley2007understanding}.
 \item \textbf{WM (Working Memory):} Participants were asked to complete two-back working memory tasks and zero-back tasks with four different types of image stimuli (places, tools, faces or body parts).
\end{enumerate}

\begin{table}[htbp]
 \centering
  \caption{Number of scans per session and its duration (min:sec)}
 {\small
  \begin{tabular}{lccccccc}
   \hline
    & Emotion & Gambling & Language & Motor & Relational & Social & WM\\
   \hline
   Scans & 176 & 253 & 316 & 284 & 232 & 274 & 405  \\
   Duration & 2:16 & 3:12 & 3:57 & 3:34 & 2:56 & 3:27 & 5:01   \\
   \hline
  \end{tabular}
  \label{tab:1}
 }
\end{table}

\subsection{Deep neural networks}
\begin{figure}[htpb]
\begin{center}
\includegraphics[width=1.05\columnwidth]{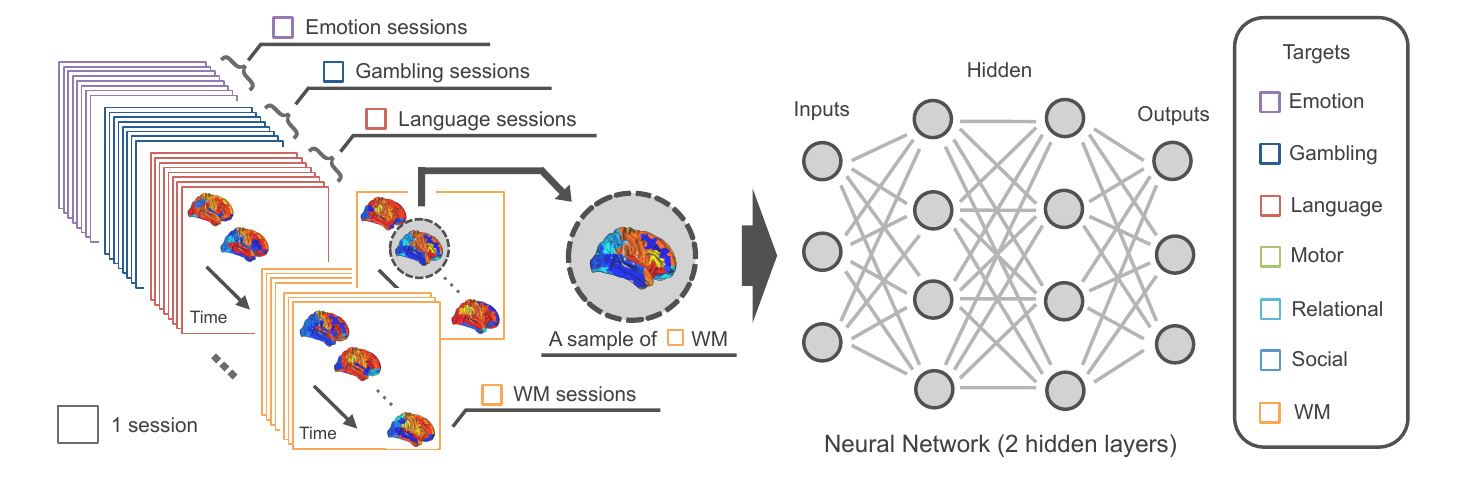}
\caption{\label{fig:dnn}
We trained DNNs with the input being the fMRI scans and the output being their labeled task classes.}
\end{center}
\end{figure}

We trained a deep neural network (DNN) with the input being the fMRI signals over aROIs and the output being their labeled task classes, i.e., the category of cognitive task performed by the participants.
Prior to the training step, all fMRI scans were categorized into seven task classes, completely disregarding the time order. The weight parameters of the DNN were then trained to optimize the probability of successfully classifying the fMRI scans into the seven task categories (Fig.\,\ref{fig:dnn}).

We trained feed-forward neural networks (i.e., DNNs) with $L$ hidden layers with stochastic gradient descent (SGD) with dropout \citep{Hinton2012}.
The internal potential of the $i$-th unit in the $l$-th hidden layer $a^{(l)}_i$ $(l = 1, \cdots, L)$ is given as a weighted summation of its inputs:
\begin{equation}
 a^{(l)}_i = \sum_{j=1}^{n_{l-1}} w^{(l)}_{ij} z^{(l-1)}_j + b^{(l)}_i,
\end{equation}
where $w^{(l)}_{ij}$ and $b^{(l)}_i$ are a weight and a bias, respectively.
Here, $n_l$ is the number of units in the $l$-th hidden layer, which was set at $n_l = 500$ for any $l > 0$.
We define $\bvec{z}^{(0)}$ as the input vector $\bvec{x}$ to the network.
This forces $n_0$ to be equal to the input's dimensionality $d (=116)$.
We denote $\bvec{z}^{(l)} = \left(z^{(l)}_1, \cdots, z^{(l)}_{n_l}\right)$ as the outputs of the $l$-th hidden layer.
These outputs are given by applying a nonlinear activation function $h$ to the internal potential as
\begin{equation}
 z^{(l)}_i = h(a^{(l)}_i)_{.}
\end{equation}
Here ReLU \citep{jarrett2009best}, a piecewise linear function $max(0, x)$, was used for the activation function $h$.
ReLU as a choice of the activation function has a couple of advantages; its piecewise linearity can save the computational cost to calculate its derivative, and its non-saturating character in the positive domain prevents the learning algorithm from halting due to gradient vanishing of nonlinear activation functions.
The last hidden layer was connected to the softmax (output) layer, so that the output from the $k$-th unit of the output layer can be interpreted as the posterior probability of class $k$, given by
\begin{equation}
 P(Y = k \,|\, \bvec{x}, \bvec{W})
=
\frac{\exp\left(\sum_{j=1}^{n_L}w_{kj} z^{(L)}_j + b_k\right)}
{\sum_{k^{'}=1}^{K}\exp\left(\sum_{j=1}^{n_L}w_{k^{'}j} z^{(L)}_j + b_{k^{'}}\right)},
\end{equation}
where $K (=7)$ is the number of classes, and $\bvec{W}$ denotes all the parameters (weights and biases) of the whole network.
Here, $Y$ is a random variable signifying the class to which $\bvec{x}$ belongs.

We used a negative log-likelihood as the cost function of the learning
\begin{equation}
  L(\bvec{W}) = - \sum_{n=1}^{N} \log P(Y_n = t_n \,|\, \bvec{x}_n, \bvec{W}),
\end{equation}
where $\left\{(\bvec{x}_1, t_1), \cdots, (\bvec{x}_N, t_N)\right\}$ constituted the given dataset.
Here $t \in \left\{1, \cdots, K\right\}$ denotes a class label.
To minimize the above cost function, minibatch stochastic gradient descent (MSGD) was introduced so that SGD was performed every 100 samples:
\begin{align} 
\bvec{W}_{t} &= \bvec{W}_{t-1} - \eta_{t} \left. \frac{\partial
L'(\bvec{W})}{\partial \bvec{W}} \right|_{\bvec{W}_{t-1}},
\end{align}
where $L'$ is the cost function for the cached subset of 100 samples in the minibatch, and $\eta_{t}$ is the learning rate.
For our case we adopted a constant rate $\eta_{t} = \eta_{0}$, and this value was set at either of $\{0.1, 1.0\}$ that yielded better result for the validation dataset (see Section\,\ref{sec:decoding_setting}).
Each weight of hidden layers was initialized at a small value randomly sampled from a zero-mean normal distribution with the standard deviation of $0.01$, and weights of softmax layer and biases were initialized to zero.
Early stopping was also adopted; if the decoding accuracy for the validation dataset did not increase for $100$ learning epochs, then learning was terminated.

To further avoid over-fitting, we used the dropout technique \citep{Hinton2012}.
During the training, the activity $z^{(l)}_i$ was randomly replaced by $0$ with probability $p$.
We set $p = 0.5$ for hidden units and $0.2$ for inputs.
This drop-out procedure plays a role of regularization
and is expected to prevent the decoder from acquiring subject-specific features.
When testing the trained neural network,
all the nodes were activated, and their weights were multiplied by $1-p$.
This is to make the mean activity level of each network element consistent
between the training phase and the test phase.

\subsection{Subject-transfer decoding}
\label{sec:decoding_setting}
To examine the subject-transferability of our decoder's architecture, we selected hundred individuals from all $499$ subjects who (1) are unrelated with each other and (2) successfully completed all seven cognitive tasks twice.
Let $D$ be the dataset of these $100$ subjects.
We then executed a \textit{leave-10-subjects-out} (or \textit{10-folds}) cross calidation to the dataset $D$.
To be more specific, the $D$ was partitioned into a test dataset of $10$ subjects, a validation dataset of $10$ subjects, and a training dataset of $80$ subjects without any overlap.
We trained our DNN with the training dataset, while using the validation dataset to determine the hyper parameters and to perform early stopping.
We then tested the decoding accuracy of trained DNNs over the test dataset.
We repeated this cycle $10$ times, choosing different test and validation datasets in each iteration.
In order to examine how the size of the dataset influences the decoding accuracy, we conducted the above experiments with different size of training dataset ($10 \sim 80$ subjects) without changing the test dataset and the validation dataset.

\subsection{Analysis for trained decoder}
The purpose of the brain decoder is here to classify the brain activities into seven categories.
The result of the classification itself, however, does not necessarily give informatoin of neuroscientific bases behind the classification.
One might therefore wish to investigate the decoder in an attempt to learn the signature that characterizes each target class.
This approach relies on the philosophy of ``knowledge discovery,'' and one may interpret these signatures acquired by the decoder as the decoder's knowledge.
Any association between the knowledge of the brain decoder and the knowledge of the neuroscience can help assess the reason why the decoder performs well/terrible, and in some case help understand the neural bases useful in brain decoding.
In the case of linear decoders, the weight visualization is often used for this purpose \citep{miyawaki2008visual,abraham2014machine}.
Such visualization can be inappropriate for decoders with a nonlinear and hierarchical architecture like DNN because the middle layer will mask the direct relation between the input and output.
A well known alternative to the weight visualization in such cases is the sensitivity analysis \citep{Zurada1994,Zurada1997,Kjems2002}, which computes the expected sensitivity of the classifier's output (posterior probability of the successful classification) with respect to the perturbation in the input.
In this study, we applied principal sensitivity analysis (PSA)
introduced by the authors \citep{Koyamada2014}, a PCA like extension of the sensitivity analysis, to our DNN-based decoder.
PSA distinguishes itself from the ordinary sensitivity analysis in that it can identify the direction in the input space to which the classifier is most sensitive.
It can also decompose the input space into the classifier-sensitive spaces and rank them in order of sensitivity.
In next two subsections, we briefly describe the sensitivity analysis and PSA.

\subsubsection{Sensitivity analysis}
Let $f_k(\bvec{x}):\RR^{\mathrm{d}} \rightarrow \RR$ be the logarithm of the output from the $k$-th unit in the final layer, namely
\begin{equation}
\label{eq:f_k}
f_k(\bvec{x}) := \log P(Y = k \,|\, \bvec{x}, \bvec{W}),
\end{equation}
where $\bvec{W}$ is the parameters of the trained decoder.  For simplicity, we would omit index $k$ in the rest of this section.
The sensitivity of $f$ with respect to the $i$-th input feature is defined by
\begin{equation}
\label{eq:original_sensitivity}
s_i := E_q \left[ \left(\frac{\partial f(\bvec{x})}{\partial x_i} \right)^2 \right],
\end{equation}
where $q$ is the true input distribution.
In actual implementation, the expectation \eqref{eq:original_sensitivity} is computed with respect to the empirical distribution of the test dataset.
\citet{Kjems2002} defined the vector
\begin{equation}
\label{eq:classical_def}
\bvec{s} := \left(s_1, \dots, s_d \right)
\end{equation}
of these values as \textbf{sensitivity map} over the set of input features.
This sensitivity map will give us a measure for the degree of importance that the classifier puts to each input.

\subsubsection{Principal Sensitivity Analysis (PSA)}
The purpose of the PSA is  to compute the direction $\bvec{v}$ for which $f$ is most sensitive in the input space.
This amounts to solving the following optimization problem about $\bvec{v}$:
\begin{eqnarray}
\label{eq:psm}
\begin{aligned}
 & \text{maximize}
 & & s(\bvec{v})\\
 & \text{subject to}
 & & \bvec{v}^{\mathrm{T}}\bvec{v} = 1,
\end{aligned}
\end{eqnarray}
where $s(\bvec{v})$ is the sensitivity of $f$ for the direction $\bvec{v}$, given by
\begin{equation}
\label{eq:sensitivity_for_direction}
 s(\bvec{v}) := E_q\left[ \left\| \nabla_{\bvec{v}} f(\bvec{x}) \right\|_2^2 \right],
\end{equation}
where $\left\|\cdot\right\|_2$ defines the L$2$-norm.
Recall that the directional derivative is defined by
\begin{equation}
 \nabla_{\bvec{v}} f(\bvec{x}) = \bvec{v}^{\mathrm{T}} \nabla f(\bvec{x}).
\end{equation}
Because we can rewrite $s(\bvec{v})$ as $\bvec{v}^{\mathrm{T}}\bvec{K}\bvec{v}$, where $K := E_{q} \left[ \nabla f(\bvec{x}) \nabla f(\bvec{x})^{\mathrm{T}} \right]$, the optimization problem \eqref{eq:psm} equals to
\begin{eqnarray}
\label{eq:psm_rewrite}
\begin{aligned}
 & \text{maximize}
 & & \bvec{v}^{\mathrm{T}}\bvec{K}\bvec{v}\\
 & \text{subject to}
 & & \bvec{v}^{\mathrm{T}}\bvec{v} = 1.
\end{aligned}
\end{eqnarray}
The solution to this problem is simply the maximal eigenvector $\pm\bvec{v}^{*}$ of $\bvec{K}$.
\citet{Koyamada2014} defined this vector as \textbf{principal sensitivity map (PSM)} over the space of input features.
The magnitude of $v_i$ represents the extent to which $f$ is sensitive to the $i$-th input feature, and the sign of $v_i$ tells us the relative direction to which the input feature influences $f$.
Recall that, if the positive definite matrix $\bvec{K}$ is replaced by the covariance $E_q[\bvec{x}\bvec{x}^{\mathrm{T}}]$, where $\bvec{x}$ is the centered random variable, the optimization problem \eqref{eq:psm_rewrite} can be seen as the problem of solving for the first principal component of the ordinary PCA.
Since the $k$-th component of the ordinary PCA is the $k$-th dominant eigenvector of the covariance matrix, the $k$-th dominant eigenvector of $\bvec{K}$ can be called the \textbf{$k$-th principal sensitivity map ($k$-th PSM)}.
These sub-principal sensitivity maps grant us access to even richer
information that underlies the dataset through the classifier.

\section{Results}

First, we compared the decoding accuracy of the DNNs with those of other baseline methods using the dataset $D$ (see Section\,\ref{sec:decoding_setting}).
We trained three neural networks with one, two and three hidden layers, each with the output logistic regression layer.
The baseline methods investigated in this study
include logistic (softmax) regression, which corresponds to 0-hidden layer neural network and SVMs with linear kernel and RBF kernel (see Appendix A for the specification of these baseline methods).
The mean decoding accuracy and its standard deviation in the \textit{leave-10-subjects-out} cross validation are summarized in Table\,\ref{tab:decoding_accuracy}.
The decoding accuracies of the DNN decoders not only exceeded the \textit{prior} chance level (the true fraction of the largest class, $20.88$\%), but were also higher than those of the other baseline methods.
In particular, the DNN with two hidden layers exhibited the best decoding accuracy of $50.74$\%, which was  significantly higher than that of the RBF-kernel SVM ($p < 0.01 $, one-sided Welch test).
Linear methods, the logistic regression and the linear-kernel SVM,
showed poor decoding accuracies that are comparable to the chance level.
These results clearly show the advantage of nonlinear decoders over linear decoders in the subject-transfer setting, and suggest that the DNN may be more effective in extracting subject-independent features from big data than the other baseline methods.

\begin{table}
\label{tab:decoding_accuracy}
 {\small
 \caption{Sujbect-transfer decoding performance}
 \centering
  \begin{tabular}{lrr}
  \hline
  \textbf{Method} &\textbf{Architecture} & \textbf{Mean accuracy [\%] $\pm$ s.d.} \\
  \hline
  Logistic regression & $116$-$7$ & 20.81 $\pm$ 0.15 \\
  Support vector machine & Linear kernel & 20.87 $\pm$ 0.01 \\
  Support vector machine & RBF kernel & 47.97 $\pm$ 1.57 \\
  Neural network & $116$-$500$-$7$ & 48.94 $\pm$ 1.15 \\
  Neural network & $116$-$500$-$500$-$7$ & \textbf{50.74} $\pm$ 1.25 \\
  Neural network & $116$-$500$-$500$-$500$-$7$ & 50.57 $\pm$ 1.31 \\
  \hline
 \textit{Prior} chance level & & 20.88 \\ 
  \hline
 \end{tabular}
}
\end{table}

\begin{figure}[htbp]
\begin{center}
\includegraphics[width=0.98\columnwidth]{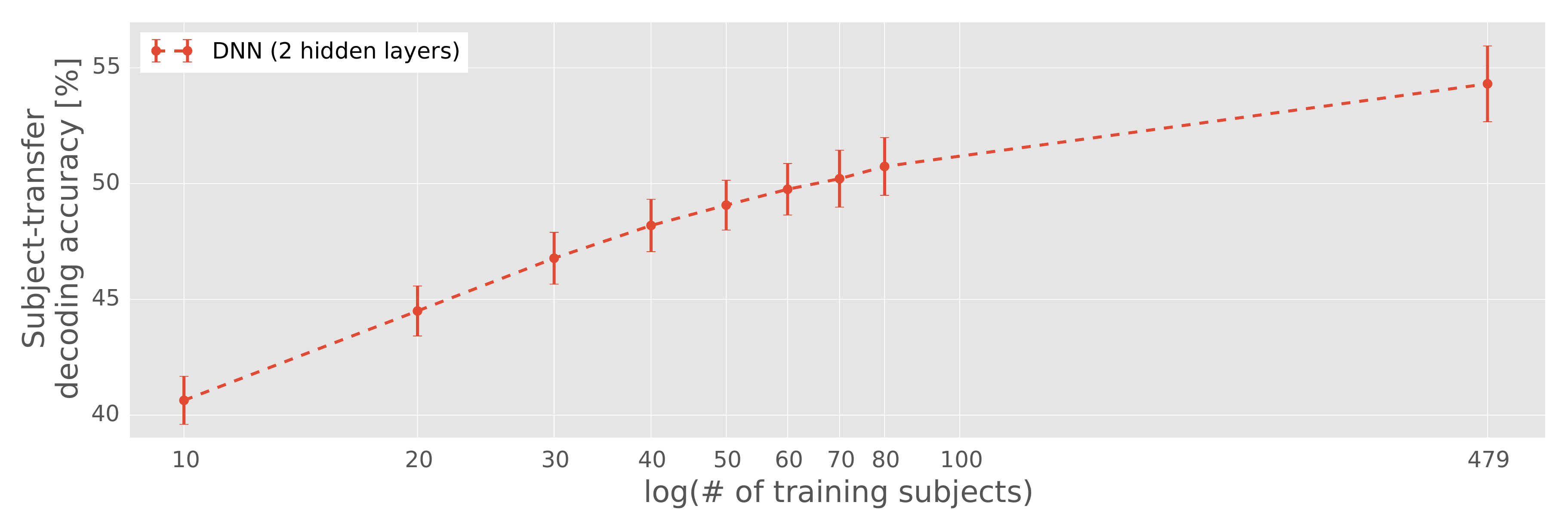}
\caption{\label{fig:change_n}
Mean decoding accuracy plotted against the number of training subjects ($M$) in log scale.
Error bars indicate s.d. across ten cross validation iterations.}
\end{center}
\end{figure}

Second, we investigated how the decoder's performance changes with the size of training dataset.
We trained the decoder with various sizes of training dataset, and plotted the decoding accuracy against the dataset size.
In this set of experiments, we employed the DNN with two hidden layers ($L = 2$),
which showed better decoding accuracy than the $L = 1$ and $L = 3$ versions over the dataset $D$.
To evaluate the performance of a DNN decoder trained with a training set of $M$ subjects,
we used the following cross validation procedure.
The setup of the cross validation is basically same as the one explained in Section\,\ref{sec:decoding_setting}.
At each iteration of the cross validation procedure, we selected $10$ subjects for the test set, $10$ subjects for the validation set, and $M$ subjects for the training set from the dataset $D$ without any overlap.
We repeated this process $10$ times, selecting $10$ entirely new subjects for the test set at each iteration.
We conducted this set of iteration procedure for $M = 10 \sim 80$.
In order to check the asymptotic trend of the decoding accuracy, we examined the $M = 479$ case as well, in which all of the $499$ subjects registered in the S500 release were used.
The results are displayed in Fig.\,\ref{fig:change_n}.
As the number of subjects in the training dataset increased from $10$ to $80$, the performance of the DNN decoder also increased, as expected.
The performance was best at $M = 479$.
We attribute this trend to the positive relationship between the size of the training dataset and the reliability of the subject-independent features captured by the DNN decoder.
This result also implies that our DNN-based subject-transfer decoding would become increasingly more practical if we can access to the brain signal databases gathered from even larger set of subjects.
DNN together with big data thus proves to be an effective approach in subject-transfer decoding.

In order to extract the discriminative features in brain activities captured by brain decoder trained with the large scale database, we conducted principal sensitivity analysis (PSA).
PSA is different from the standard sensitivity analysis in that it quantifies the \textit{combinations} of aROIs used in the decoder's classification, whereas the standard sensitivity analysis computes the independent contribution of each aROI to the decoder's decision.
The PSA is more suited for our purpose because there is a strong evidence of functional connectivity among aROIs \citep{Buckner2013, Cole2014}, and classifiers with high decoding accuracy are likely to capture coactivation patterns of aROIs.
The expectation in equation\,\eqref{eq:sensitivity_for_direction} depends on the distribution $q$ over the sample space.
In order to best approximate the true distribution of $q$, we used the empirical distribution over the dataset of all subjects except the subjects used in $D$.
Fig.\,\ref{fig:PSA}(a) shows the first, second and third PSMs for Emotion and Motor classes\footnote{See Appendix B Fig.\,\ref{fig:appendix_psa} for the PSMs of the other classes.}.
The set of PSMs presented in Fig.\,\ref{fig:PSA}(a) is one of the $10$ variants of the sets of PSMs obtained in the \textit{leave-$10$-subjects-out} cross validation over the dataset $D$.
These variants did not exhibit siginificant variation.
PSMs are superior to standard sensitibity maps in that they can describe the aROIs which act oppositely in characterizing the class.
Any pair of anatomical regions with different color assignments in Fig.\,\ref{fig:PSA} contributes to the classifier's decision in opposite direction.
Our PSA seems to imply that the information learned by our DNN-based subject-transferable decoder has some correlation with the existing knowledge of functional connectivity supported in neuroscience.
For instance,  in the 2nd PSM of Motor class and the first PSM of Social class (Fig.\,\ref{fig:PSA}),
we can identify the two sets of functional connectivitiy established in previous works, namely fronto-parietal network \citep{cole2013multi} and salience network \citep{taylor2009two}.
In the 2nd PSM of Social class (Fig.\,\ref{fig:PSA}), we can also find a component of the default mode network \citep{raichle2007default, raichle2001default} in the left hemisphere.
In addition, to quantify the similarity of PSMs, we calculated the absolute cosine similarity for each pair of the PSMs and aligned the maps by hierarchical clustering (Fig.\,\ref{fig:PSA}(b)).
For any pair of PSMs ($\bvec{v}_1, \bvec{v}_2$), the absolute cosine similarity was calculated by $\left|\langle\bvec{v}_1, \bvec{v}_2\rangle\right|$, where $\langle\cdot, \cdot\rangle$ is an inner product, because of $\|\bvec{v}\|_2 = 1$ for each PSM $\bvec{v}$.
In the similarity matrix, we can confirm two large clusters, consisting mainly of first and second PSMs, respectively.
This implies that the funcational connectivity interpretation that we
made above for the first and the second PSMs of the Social class and the
the first PSM of the Motor class applies to all the other PSMs sharing
the same cluster memberships (see also Appendix B Fig.\,\ref{fig:appendix_psa}).
On the other hand, the sub-principal PSMs, such as the third PMSs of Emotion and Motor classes, showed task-specific features.
Finally, note that many of our PSMs span large brain regions.
This suggests that the subject independent features that we extracted from the large fMRI database in our deep learning procedure are specific(common) brain-wide networks that activates during particular(all) task(s).

\begin{figure}[htpb]
\begin{center}
\includegraphics[width=0.99\columnwidth]{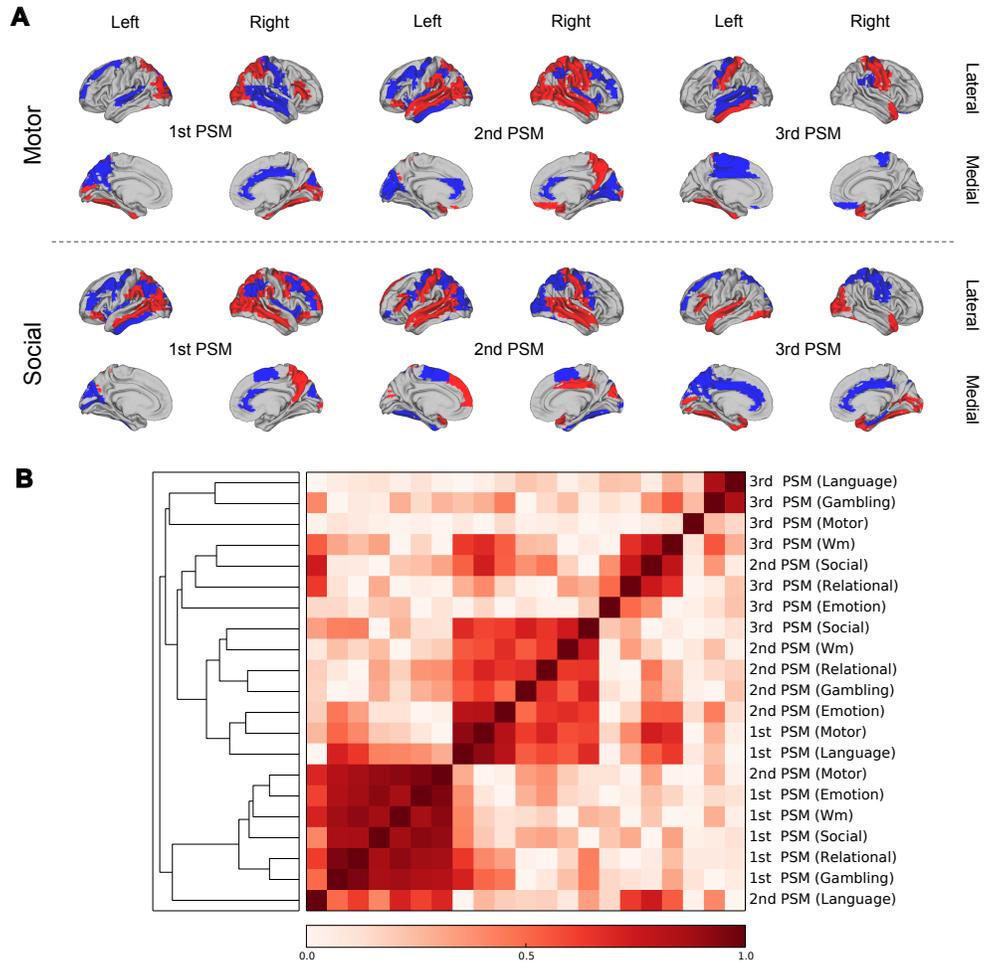}
\caption{\label{fig:PSA}
(a)
We calculated the first, second and third PSMs for each class. The PSMs of Motor and Social classes are shown here (see Appendix B Fig.\,\ref{fig:appendix_psa} for the other classes).
We recall that we intentionally omitted the class index in the formulation \eqref{eq:psm_rewrite} for simplicity,
and that PSMs are actually meant to be computed for each class separately.
In the visualization of each PSM, we exclusivly colored the ROIs with values that are at least one s.d. away from the mean.
Red and blue indicate different signs in the PSM values, i.e., the corresponding aROIs act oppositely in characterizing the class.
(b)
Similarity between maps was evaluated by cosine similarity.
We calculated the absolute value of cosine similarity (for definition, see the text) between all pairs of the first, second and third PSMs of the seven classes. Based on these similarity values, we applied a hierarchical clustering to the set of all the computed PSMs.
The similarity matrix above is based on leaf order in the dendrogram (attached to the left side of the matrix) obtained by the hierarchical clustering. Intensity of the color indicates the degree of similarity.
}
\end{center}
\end{figure}

\section{Conclusion}
In this study, we applied deep learning to a large fMRI database to classify brain activities into seven human categories.
In particular, we constructed a DNN-based brain decoder aimed at classifying the brain activities of
\textit{any arbitrary} individual.
The strength of DNN is its high ability to capture nonlinear, high dimensional features from bigdata.
In fact, the decoding accuracy of our DNN was superior to those of other baseline methods, including SVMs
and logistic regressions.
The high performance of our DNN over the dataset acquired from a large population is indicative of the ability of
our decoder to capture nonlinear discriminative features that are common to all subjects.
Interestingly, when we visualzied these universal discriminative features in the form of PSMs, we were able to find non trivial associations between the features and functionally connected networks recognized in neuroscience.
This observation suggests that the functional connectivity common to all subjects are playing important roles in characterizing task-specific brain activities.
Furthermore, our DNN-based decoder leaves some room for further improvement.
For instance, the training of DNN and its final performance depend largely on the initial parameters.
One may use our DNN as a initial model to construct a decoder that is highly tuned for a specific individual.
One can expect such decoder to utilize both subject-independent features and subject-specific features.
As another interesting extension to this research, we may add demographic features like age and sex to the model.
The flexibility of DNN architecture allows for numerous modification of the base model.
Also, the relationship between the number of training subjects and the decoding accuracy was positive.
With a bigger and more multimodal dataset and a DNN with more sophisticated architecture, one might be able to capture richer signatures that are otherwise difficult to uncover, and such signatures might inspire a novel insight into neural bases recruited in different situations in the brain.
Our results hint that advanced machine learning techniques will continue
to grow in importance in the coming era of computational neuroscience that finds its basis in the statistics of heterogenous big data.

\section*{Acknowledgement}
Data were provided by the Human Connectome Project,
WU-Minn Consortium\footnote{Principal Investigators: David Van Essen and Kamil Ugurbil;
1U54MH091657} funded by the 16 NIH Institutes and Centers that support
the NIH Blueprint for Neuroscience Research; and by the McDonnell Center
for Systems Neuroscience at Washington University.

\appendix
\section{Specifications of baseline methods}
Each version of SVM consists of seven one-versus-the-rest classifiers.
%
As for the SVMs, we used scikit-learn \citep{pedregosa2011scikit}.
Hyper-parameters were chosen to maximize the decoding accuracy over the validation dataset (see Section\,\ref{sec:decoding_setting}); we heuristically prepared nine sets of hyper-parameters for each method, and adopted the one that achieved the best decoding accuracy for the validation dataset.
The hyper-parameter for the logistic regression was the learning rate in the MSGD.
The hyper-parameter for the linear-kernel SVM was the regularization strength $C$ used in the scikit-learn.
The RBM-kernel SVM was dependent on a pair of hyper-parameters, the regularization strength $C$ and the kernel width $\gamma$.
We considered $3$ values each for $C$ and $\gamma$, and examined all nine pairs for the best choice.

\section{PSMs of the subject-transfer decoder}

\begin{figure}[htbp]
\begin{center}
\includegraphics[width=0.99\columnwidth]{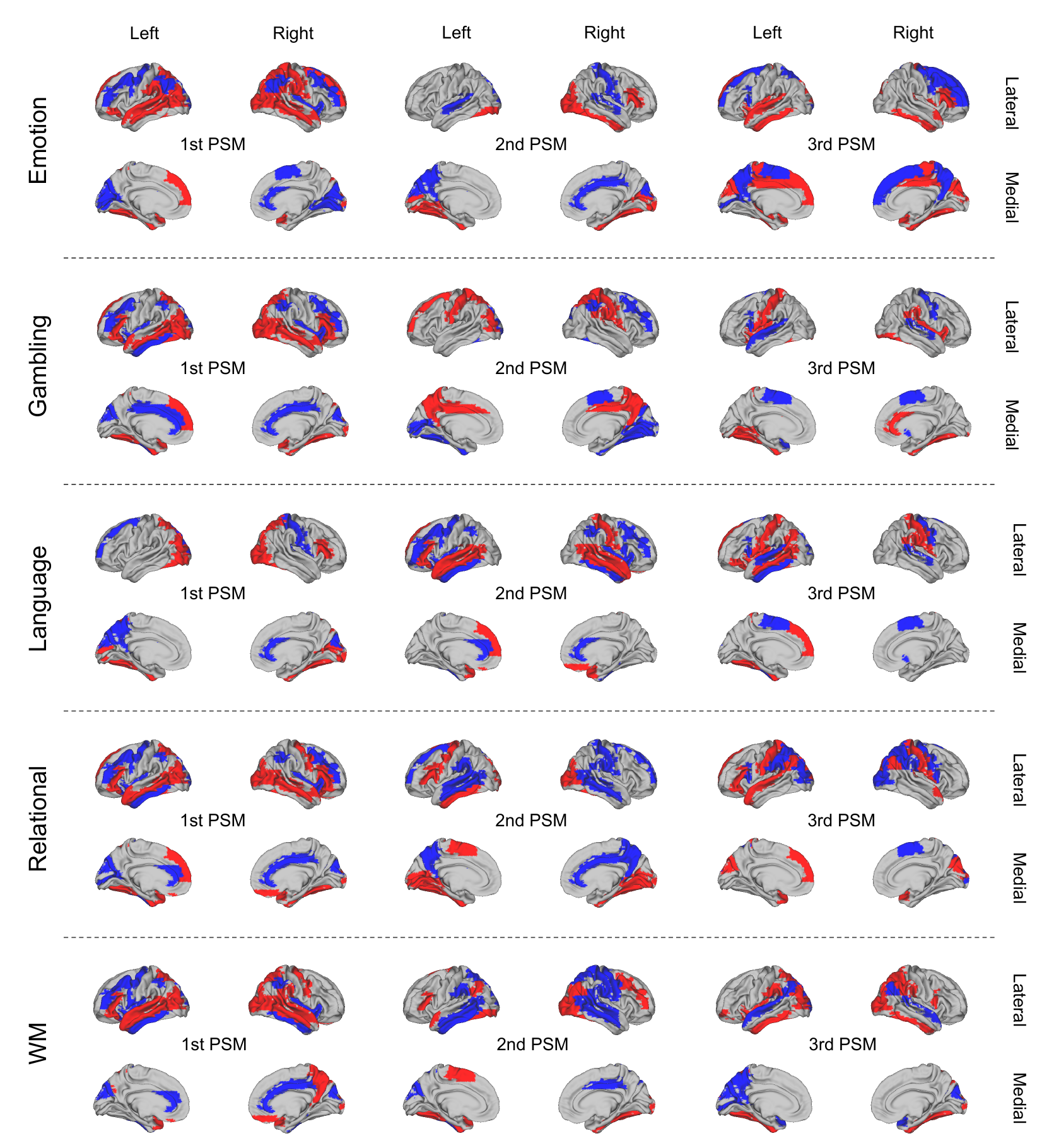}
\caption{\label{fig:appendix_psa}
PSMs of the other classes.
See the caption of Fig.\,\ref{fig:PSA} for the specification.}
\end{center}
\end{figure}